\documentclass{article}
\usepackage{spconf,amsmath, graphicx, array, book tabs, tabularray, multirow}

\usepackage{enumitem}
\setlist{nosep, leftmargin=14pt}

\usepackage{mwe} 
\usepackage{placeins}
\usepackage{hyperref}
\usepackage{multirow}
\usepackage{threeparttable}
\usepackage{adjustbox}
\usepackage{xcolor}
\usepackage{lipsum}                 
\usepackage{makecell} 

\newcommand{\todo}[1]{}
\renewcommand{\todo}[1]{{\color{red} TODO: {#1}}}

\newcommand{\TODO}[1]{}
\renewcommand{\TODO}[1]{{\color{red} TODO: {#1}}}

\title{melanoma detection with uncertainty quantification}

\name{SangHyuk Kim$^{1,2}$ \qquad Edward Gaibor$^{1}$ \qquad Brian Matejek$^{2}$ \qquad Daniel Haehn$^{1}$}

\address{
$^{1}$ Department of Computer Science, University of Massachusetts Boston, Boston, Massachusetts, USA \\
$^{2}$Computer Science Laboratory, SRI International, Arlington, Virginia, USA\
}

\let\OLDthebibliography\thebibliography
\renewcommand\thebibliography[1]{
  \OLDthebibliography{#1}
  \setlength{\parskip}{0pt}
  \setlength{\itemsep}{0pt plus 0.3ex}
}

\begin{document}
%
\maketitle
\begin{abstract}
  Early detection of melanoma is crucial for improving survival rates. Current detection tools often utilize data-driven machine learning methods but often overlook the full integration of multiple datasets. We combine publicly available datasets to enhance data diversity, allowing numerous experiments to train and evaluate various classifiers. We then calibrate them to minimize misdiagnoses by incorporating uncertainty quantification. Our experiments on benchmark datasets show accuracies of up to 93.2\% before and 97.8\% after applying uncertainty-based rejection, leading to a reduction in misdiagnoses by over 40.5\%. Our code and data are publicly available\footnote{Our GitHub repository, \url{https://mpsych.org/melanoma}}, and a web-based interface\footnote{\label{footnote:webinterface}Our Web interface, \url{https://mpsych.github.io/melanoma/}} for quick melanoma detection of user-supplied images is also provided.
  
  \end{abstract}


%
%

\section{Introduction}
\label{sec:intro}

Melanoma is a severe skin cancer responsible for around 55,500 deaths annually~\cite{melanomadeath}. Deep Neural Networks (DNNs) are effective in melanoma detection, but their evaluation lacks standardization, complicating comparisons. For example, DNN-based melanoma classification using segmented features~\cite{using16_17} has shown promising results on the ISIC'16~\cite{isic2016} and ISIC'17~\cite{isic2017} datasets. Transfer learning methods~\cite{using18_ph2} have also achieved high accuracy on ISIC'18~\cite{isic2018}. Additionally, Vision Transformers~\cite{mypaperref} have demonstrated exceptional accuracy on a private dataset.
However, many studies operate under inconsistent testing conditions.


Melanoma lesions exhibit significant variability, making detection challenging in real-world settings. While DNNs require extensive datasets for accuracy, only a few studies
have integrated multiple datasets within a unified framework. Additionally, although DNN predictions can be accurate, they may be confidently incorrect, underscoring the need for reliable diagnostic methods.

\begin{figure*}[hbtp]
    \centering
    \includegraphics[width=1\textwidth]{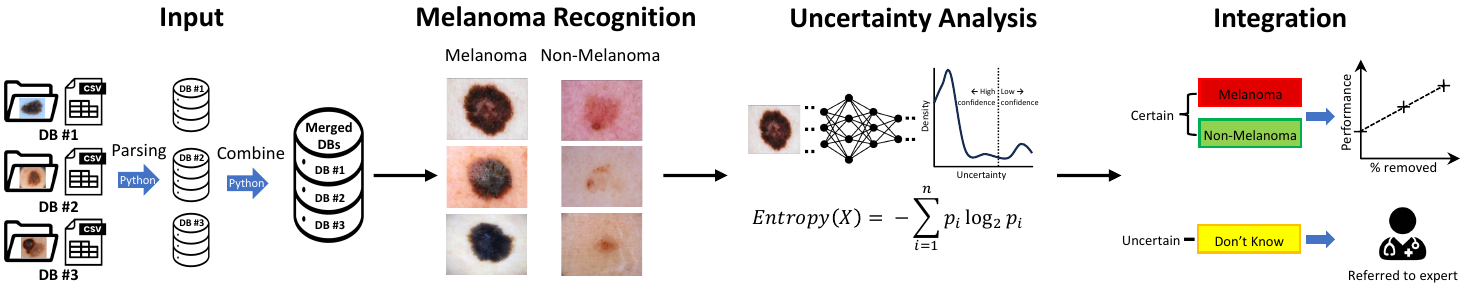} 
    \caption{
    Our framework includes aggregating diverse datasets, optimizing predictions through 1,296 experiments with 54 combinations of combined datasets and 24 CNNs, filtering uncertain predictions with entropy scores, referring ambiguous cases for human evaluation, and enhancing classification performance.
    }
    \label{fig:overview}
\end{figure*}

We develop a framework for consistent experimentation with datasets and DNNs to minimize misdiagnoses. It includes four modules: Input, Melanoma Recognition, Uncertainty Analysis, and Integration, as shown in Fig.~\ref{fig:overview}.
Our methodology involves importing multiple datasets, uniform preprocessing, and evaluating classifiers in classification, calibration, and uncertainty scores. We filter out uncertain predictions to enhance the classification and calibration performance of the remaining samples. Ultimately, we aim to develop reliable classifiers for clinical use and help researchers identify optimal dataset and classifier combinations. 


\section{Methods}
\label{sec:format}



\subsection{Input}
\label{subsec:input}
Most public melanoma datasets include images and associated metadata about skin lesions.
Image sizes range from 147x147 to 3096x3096 pixels, with class counts from 2 to 9, consistently including melanoma labels.
These datasets vary in structure—some use a single folder with a separate CSV for labels and splits, while others organize by folders or categories.
We parse these diverse structures to load them into compatible environments and create merged datasets with a common interface.
 


\subsection{Melanoma Recognition}
\label{subsec:mel_rec}
\textbf{Classification.} 
We use a unified framework for preprocessing, augmentation, training, validation, and inference, enabling direct performance comparisons across experiments. This approach combines datasets with DNN models and applies standard metrics like precision, sensitivity, specificity, F1, accuracy, and AUC-ROC.
The softmax classifier categorizes results as either ``Melanoma'' or ``Non-Melanoma''.

\noindent
\textbf{Calibration.} 
We evaluate our models' reliability using Expected Calibration Error (ECE)~\cite{guo2017calibration} and the Brier score~\cite{brier1950verification}, analyzing how different data combinations affect these metrics to ensure model confidence aligns with actual outcomes.



\begin{figure}[th!]
\centering
\includegraphics[width=0.4\textwidth]{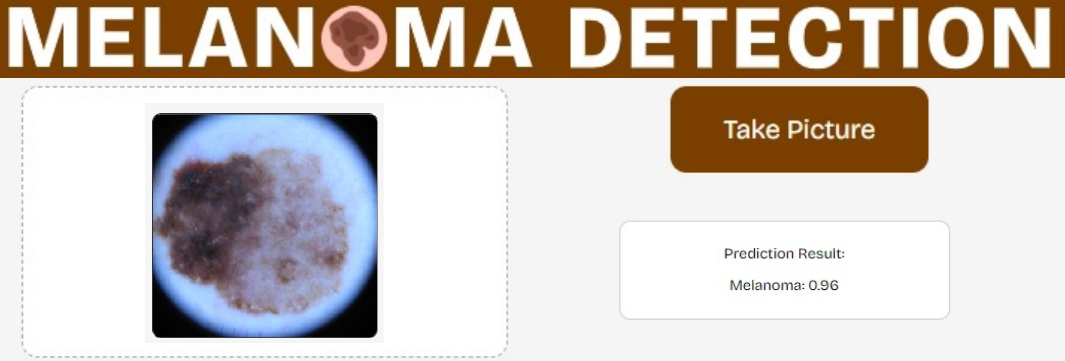}
\caption{We developed a web application that allows users to estimate a melanoma risk score of user-supplied images using web-based inference without upload (edge computing).}
\label{fig:standalone_web}
\end{figure}

\subsection{Uncertainty Analysis}
\label{subsec:uncertainty}
Our third module estimates neural network uncertainties using Shannon's entropy~\cite{shannon}. This method directly assesses the model's confidence in its predictions while considering the entire distribution of predicted probabilities, making it sensitive to model indecision.
We filter out uncertain predictions to avoid confident false diagnoses from the softmax classifier. The uncertainty scores are defined by: \\
\begin{equation}
H(X) = - \sum_{i=1}^{n} p_i \log(p_i)
\label{eq:shannon}
\end{equation}

Here, \(p_i\) is the probability of the \(i\)-th class, and \(\log(p_i)\) measures the associated uncertainty. By multiplying \(p_i\) with \(\log(p_i)\) and summing across all classes, we assess the model's uncertainty $H(X)$ within [0, 1]. Lower values indicate higher confidence, while higher values signify greater uncertainty.

\subsection{Integration}
\label{subsec:integration}
We classify images into an ``Uncertain'' class by combining neural network predictions with uncertainty scores and rejecting low-confidence predictions with high entropy. This enhances accuracy by retaining only confident predictions and excluding uncertain cases from evaluation.

\subsection{Web Application}
\label{subsec:melad}


We present a web application\textsuperscript{\ref{footnote:webinterface}} for melanoma detection that utilizes a MeshNet~\cite{meshnet} architecture with volumetric dilated convolutions to reduce parameters for web deployment (see Fig.~\ref{fig:standalone_web}). It allows users to process local images from their devices without upload and detect melanoma within seconds using client-side inference. The model is trained on 10 combined datasets~\cite{isic2016,isic2017,isic2018,isic2019,isic2020,ph2,7pointcriteria,padufes20,mednode,kagglemb}.

\section{Experimental Setup}
\label{sec:experiments}

\subsection{Datasets and Preprocessing}
\label{subsec:datasets}
We utilize publicly available skin disease datasets~\cite{isic2016,isic2017,isic2018,isic2019,isic2020,ph2,7pointcriteria,padufes20,mednode,kagglemb} and evaluate our experiments using provided test sets~\cite{isic2017,isic2018,7pointcriteria,kagglemb}. When a validation set is not provided, we split the training set into an 80:20 ratio for creating a validation set.  These datasets include a “Melanoma” label, and we convert other labels to “Non-Melanoma” for binary classification. To address the high class imbalance leaning towards “Non-Melanoma,” we apply oversampling and data augmentation techniques, such as random resized crop, color jitter, zooming, and random flip, to prevent overfitting.
We also create combined datasets using up to 10 datasets to train and evaluate various DNNs on classification and calibration, assessing the impact of added data on these metrics.


\begin{table*}[h!]{
\caption{Classification performance of our top-performing models in our curated benchmarks.}
\centering

\begin{adjustbox}{width=1\textwidth}{
\begin{threeparttable}{
\normalsize
\begin{tabular}{c c c c c c c c c c}

\hline
\multirow{2}{*}{Test set} &
\multirow{2}{*}{Network} &
\multirow{2}{*}{Train sets} &
\multicolumn{7}{|c}{Before rejection / After rejection} \\ \multicolumn{3}{c}{} &

\multicolumn{1}{|c}{Precision ($\uparrow$)} 
& \multicolumn{1}{c}{Specificity ($\uparrow$)}
& \multicolumn{1}{c}{Sensitivity ($\uparrow$)} 
& \multicolumn{1}{c}{F-1 ($\uparrow$)} 
& \multicolumn{1}{c}{Accuracy ($\uparrow$)} 
& \multicolumn{1}{c}{AUC-ROC ($\uparrow$)}
& \multicolumn{1}{c}{Threshold}\\
\hline

\multirow{3}{*}{ISIC2017~\cite{isic2017}}  &
\multicolumn{1}{c}{DenseNet201~\cite{densenet}} & 
\multicolumn{1}{c}{[A--E,G,I,J]} &
\multicolumn{1}{|c}{\textbf{83.8\%} / 85.3\%} & 
\multicolumn{1}{c}{91.5\% / 91.8\%} &
\multicolumn{1}{c}{86.3\% / 92.2\%} & 
\multicolumn{1}{c}{\textbf{86.0\%} / 88.0\%} & 
\multicolumn{1}{c}{\textbf{90.5\%} / 91.9\%} &
\multicolumn{1}{c}{94.5\% / 95.7\%} 
& \multicolumn{1}{c}{0.1}
\\

& \multicolumn{1}{c}{DenseNet201~\cite{densenet}}       & 
\multicolumn{1}{c}{[A--F,H]}       &
\multicolumn{1}{|c}{82.3\% / \textbf{92.8\%}}       & 
\multicolumn{1}{c}{90.1\% / \textbf{96.7\%}}       & 
\multicolumn{1}{c}{\textbf{87.2\%} / \textbf{94.5\%}}          &
\multicolumn{1}{c}{84.8\% / \textbf{94.1\%}}          &
\multicolumn{1}{c}{89.5\% / \textbf{96.3\%}} &
\multicolumn{1}{c}{\textbf{95.4\%} / \textbf{96.3\%}}
& \multicolumn{1}{c}{0.2} \\

& \multicolumn{1}{c}{ResNet152~\cite{resnet}} & 
\multicolumn{1}{c}{[A-D,I,J]}       & 
\multicolumn{1}{|c}{82.0\% / 87.0\%}       & 
\multicolumn{1}{c}{\textbf{91.7\%} / 93.9\%}         & 
\multicolumn{1}{c}{77.8\% / 89.6\%}         &
\multicolumn{1}{c}{83.2\% / 89.1\%}         &
\multicolumn{1}{c}{89.0\% / 93.1\%} &
\multicolumn{1}{c}{93.9\% / 94.4\%} & \multicolumn{1}{c}{0.2} \\
\hline
\multirow{3}{*}{ISIC2018~\cite{isic2018}}  & \multicolumn{1}{c}{ResNet152~\cite{resnet}} & \multicolumn{1}{c}{[D]} & 
\multicolumn{1}{|c}{75.0\% / 81.1\%}& 
\multicolumn{1}{c}{93.1\% / 96.4\%} & 
\multicolumn{1}{c}{\textbf{65.5\%} / 70.9\%} & 
\multicolumn{1}{c}{\textbf{77.0\%} / 82.3\%} & 
\multicolumn{1}{c}{89.9\% / 94.2\%} &
\multicolumn{1}{c}{91.0\% / 92.7\%} & \multicolumn{1}{c}{0.15} \\

& \multicolumn{1}{c}{ResNet152~\cite{resnet}} & 
\multicolumn{1}{c}{[A,C--E]}       &
\multicolumn{1}{|c}{\textbf{75.2\%} / 86.5\%}      & 
\multicolumn{1}{c}{93.4\% / 97.9\%}          & 
\multicolumn{1}{c}{63.7\% / \textbf{71.7\%}}          &
\multicolumn{1}{c}{76.7\% / \textbf{86.5\%}}          &
\multicolumn{1}{c}{\textbf{90.0\%} / 95.8\%} &
\multicolumn{1}{c}{91.1\% / 94.5\%} & \multicolumn{1}{c}{0.2} \\
                                  
& \multicolumn{1}{c}{ResNet152~\cite{resnet}} & 
\multicolumn{1}{c}{[A--D,I,J]}   & 
\multicolumn{1}{|c}{74.6\% / \textbf{88.5\%}}       & 
\multicolumn{1}{c}{\textbf{94.3\%} / \textbf{98.5\%}}         & 
\multicolumn{1}{c}{61.4\% / 67.7\%}         &
\multicolumn{1}{c}{75.9\% / 85.5\%}         &
\multicolumn{1}{c}{89.7\% / \textbf{96.0\%}} &
\multicolumn{1}{c}{\textbf{91.7\%} / \textbf{94.6\%}}
& \multicolumn{1}{c}{0.2} \\
\hline
\multirow{3}{*}{7-point criteria~\cite{7pointcriteria}}  & \multicolumn{1}{c}{ResNet152~\cite{resnet}} & 
\multicolumn{1}{c}{[A--E,G]} & 
\multicolumn{1}{|c}{\textbf{78.9\%} / \textbf{88.8\%}}  & 
\multicolumn{1}{c}{\textbf{90.1\%} / \textbf{97.1\%}} & 
\multicolumn{1}{c}{65.3\% / 65.8\%} & 
\multicolumn{1}{c}{\textbf{78.2\%} / \textbf{84.3\%}} & 
\multicolumn{1}{c}{\textbf{83.8\%} / \textbf{89.6\%}} &
\multicolumn{1}{c}{85.0\% / \textbf{86.6\%}}
& \multicolumn{1}{c}{0.2} \\

& \multicolumn{1}{c}{ResNet152~\cite{resnet}}      & 
\multicolumn{1}{c}{[A,C--E]}       &
\multicolumn{1}{|c}{77.5\% / 82.3\%}       & 
\multicolumn{1}{c}{89.1\% / 92.6\%}   & 
\multicolumn{1}{c}{64.4\% / \textbf{67.8\%}}   &
\multicolumn{1}{c}{77.0\% / 81.1\%}   &
\multicolumn{1}{c}{82.8\% / 86.5\%} &
\multicolumn{1}{c}{84.6\% / 85.4\%} & \multicolumn{1}{c}{0.1} \\
                                  
& \multicolumn{1}{c}{ResNet152~\cite{resnet}}       & 
\multicolumn{1}{c}{[A--E]}       &
\multicolumn{1}{|c}{77.4\% / 84.6\%}       & 
\multicolumn{1}{c}{88.1\% / 94.3\%}         & 
\multicolumn{1}{c}{\textbf{67.3\%} / 67.0\%}         &
\multicolumn{1}{c}{77.5\% / 82.4\%}         &
\multicolumn{1}{c}{82.8\% / 87.5\%} &
\multicolumn{1}{c}{\textbf{85.1\%} / 86.3\%}
& \multicolumn{1}{c}{0.11} \\
\hline
\multirow{3}{*}{Kaggle~\cite{kagglemb}}  & \multicolumn{1}{c}{DenseNet201~\cite{densenet}} & 
\multicolumn{1}{c}{[A--E,G,I,J]} & 
\multicolumn{1}{|c}{91.2\% / 94.1\%}  & 
\multicolumn{1}{c}{85.6\% / 88.7\%} & 
\multicolumn{1}{c}{97.3\% / 99.3\%} &
\multicolumn{1}{c}{90.9\% / 93.7\%} &
\multicolumn{1}{c}{90.9\% / 93.8\%} &
\multicolumn{1}{c}{98.1\% / 98.5\%} & \multicolumn{1}{c}{0.08} \\

& \multicolumn{1}{c}{DenseNet201~\cite{densenet}}       & 
\multicolumn{1}{c}{[A--D,I,J]}       &
\multicolumn{1}{|c}{\textbf{93.2\%} / \textbf{97.7\%}}       & 
\multicolumn{1}{c}{\textbf{89.2\%} / \textbf{96.1\%}}   & 
\multicolumn{1}{c}{98.0\% / 98.0\%}   &
\multicolumn{1}{c}{\textbf{93.2\%} / \textbf{97.8\%}}   &
\multicolumn{1}{c}{\textbf{93.2\%} / \textbf{97.8\%}} &
\multicolumn{1}{c}{98.1\% / 99.1\%} & \multicolumn{1}{c}{0.19} \\
                                  
& \multicolumn{1}{c}{ResNet101~\cite{resnet}}       & 
\multicolumn{1}{c}{[A--G,I,J]} & 
\multicolumn{1}{|c}{92.5\% / 97.6\%}       & 
\multicolumn{1}{c}{87.0\% / 95.7\%}         & 
\multicolumn{1}{c}{\textbf{98.7\%} / \textbf{99.6\%}}         &
\multicolumn{1}{c}{92.3\% / 97.6\%}         &
\multicolumn{1}{c}{92.3\% / 97.6\%} &
\multicolumn{1}{c}{\textbf{98.5\%} / \textbf{99.3\%}}
& \multicolumn{1}{c}{0.18} \\

\hline
\end{tabular}
            \textbf{Notes:} A: ISIC'16 B: ISIC'17 C: ISIC'18 D: ISIC'19 E: ISIC'20 F: 7-point criteria G: PH2 H: PAD\_UFES\_20 I: MEDNODE J: Kaggle

}\end{threeparttable}
}
\end{adjustbox}
\label{table:performance}

}\end{table*}

\subsection{Melanoma Recognition}
\label{subsec:exp_mel_rec}
We use various open-source models in \texttt{PyTorch}, including ResNet~\cite{resnet}, DenseNet~\cite{densenet}, VGG~\cite{vgg}, and EfficientNet~\cite{efficientnet} with ImageNet~\cite{imagenet} pre-trained weights, resizing images to 224x224 pixels.
We freeze the 2D convolutional layers and modify the final output layer for binary classification. Based on our empirical experiments, we train the models for a minimum of 30 epochs, gradually increasing to a maximum of 50 epochs when all ten datasets are combined. The training uses Stochastic Gradient Descent (SGD)~\cite{SGD} with a learning rate of 0.001, momentum of 0.9, and a decay factor of 0.1 every 7 epochs, employing cross-entropy as the loss function.

\begin{figure}[h!]
\centering
\includegraphics[width=0.85\columnwidth]{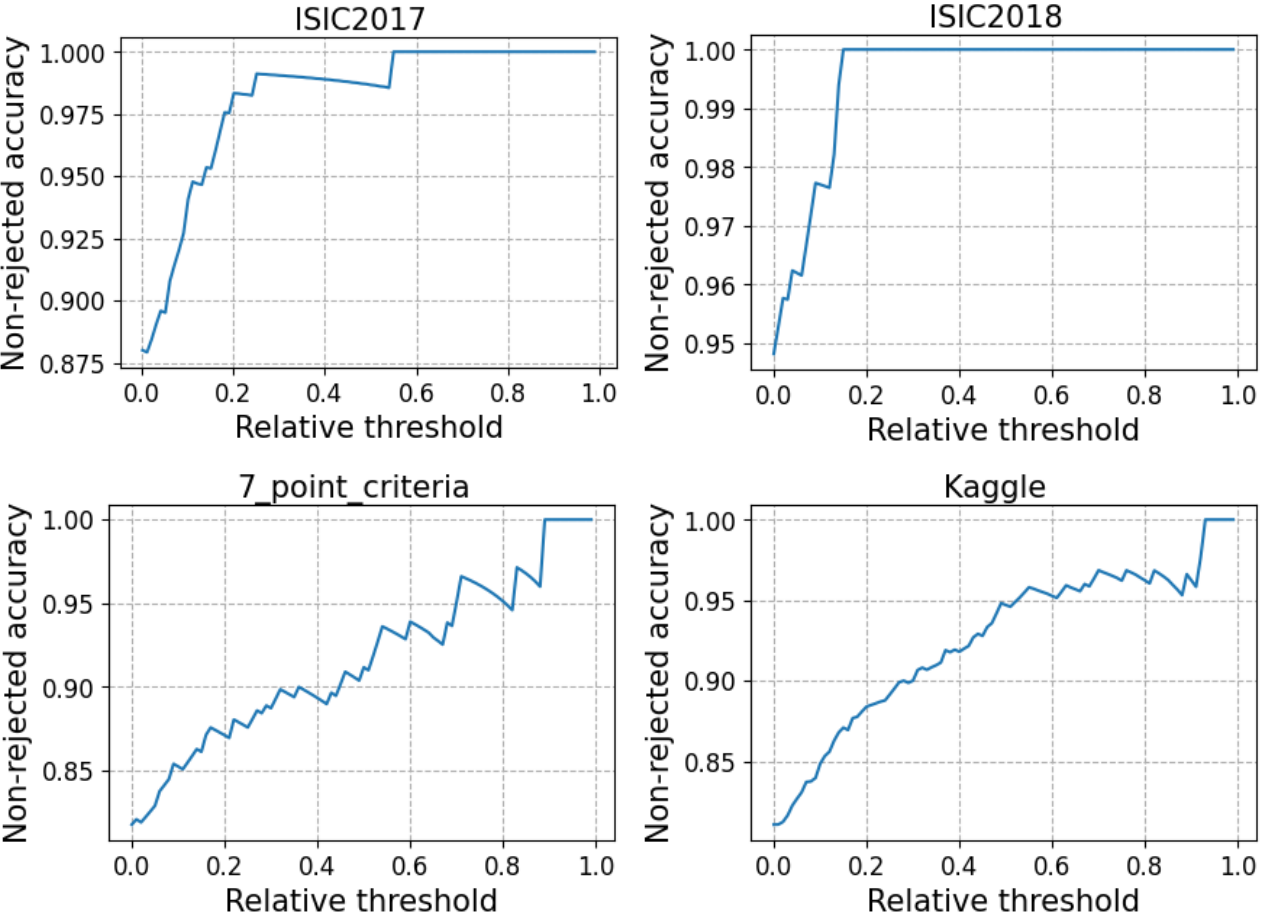}
\caption{Accuracy improves with varying rejection thresholds on validation sets, commonly in the range of 0-0.2.}
\label{fig:thresholds}
\end{figure}

\subsection{Rejection by Uncertainty}
\label{subsec:exp_uncertainty}
We perform uncertainty quantification on the predictions from trained models. 
ECE is defined as: \\
\begin{equation}
\text{ECE} = \sum_{m=1}^M \frac{|B_m|}{n} \left| \text{acc}(B_m) - \text{conf}(B_m) \right|
\end{equation}
\noindent
where \( M \) is the number of bins,
    \( B_m \) is the set of samples in the \( m \)-th bin,
    \( \text{acc}(B_m) \) and \( \text{conf}(B_m) \) are the accuracy and the average confidence 
    in the \( m \)-th bin,
     and \( n \) is the total number of samples. We use 10 bins for a finer granularity. The Brier Score is defined as: \\
\begin{equation}
\text{Brier Score} = \frac{1}{n} \sum_{i=1}^{n} (\hat{p}(x_i) - y_i)^2
\end{equation}
\noindent
where \( \hat{p}(x_i) \) is the predicted probability for sample \( x_i \), \( y_i \) is the true label, and \( n \) is the total number of samples. Both metrics range within [0, 1], with 0 indicating optimal calibration.
We find that rejecting up to 20\% of low-confidence predictions in validation sets improves model accuracy, as shown in Fig.~\ref{fig:thresholds}. We calculate ECE and Brier scores while varying the threshold to minimize their average for each model. After this dynamic rejection, we re-evaluate the remaining samples and compare them to the originals on classification and calibration metrics.


\begin{figure}[h!]
\centering
\includegraphics[width=0.95\columnwidth]{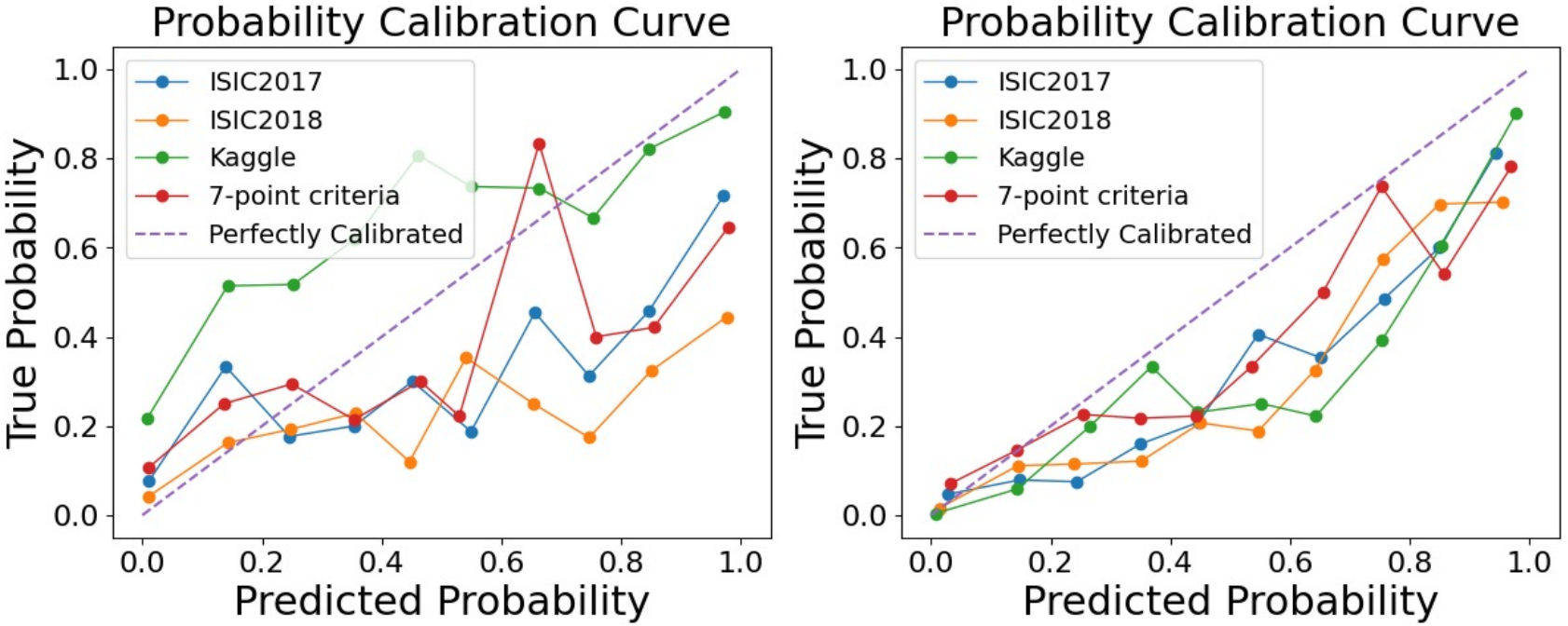}
\caption{Calibration curves of ResNet50 as a demonstration. \textbf{Left}: Single dataset (ISIC'20). \textbf{Right}: Combined datasets (ISIC'16, ISIC'17, ISIC'18, MEDNODE, Kaggle).
}
\label{fig:calibration_comparison}
\end{figure}

\section{Results}
\label{sec:results}

\textbf{Performance Before Rejection.}
 Classification performance is ranked by precision, following the ISIC challenge~\cite{isic2016}.
DenseNet201 and ResNet152 achieve the best results, with precisions of 83.8\%, 75.2\%, 78.9\%, and 93.7\% across the ISIC'17~\cite{isic2017}, ISIC'18~\cite{isic2018}, the 7-point criteria~\cite{7pointcriteria}, and Kaggle datasets~\cite{kagglemb} as shown in Table~\ref{table:performance}.
Combining datasets enhances performance; for instance, as shown in Table~\ref{table:leaderboard}, EfficientNetB1 achieves a top AUC-ROC of 90.7\% in SIIM-ISIC challenge~\cite{siim-isic-melanoma-classification} when trained on seven datasets. Additionally, models trained on combined datasets align more closely with the ideal calibration curve, represented by \( y=x \) (Fig.~\ref{fig:calibration_comparison}).
\begin{table}[t!]{
\caption{Leaderboard scores for our top-performing models on the SIIM-ISIC challenge~\cite{siim-isic-melanoma-classification}.}
\centering

\begin{adjustbox}{width=1\columnwidth}{
\begin{threeparttable}{
\normalsize
\begin{tabular}{c c c }

\hline
\multirow{1}{*}{Network} & \multirow{1}{*}{Trainsets} & \multirow{1}{*}{Private Score (AUC-ROC)} \\
\hline

\multicolumn{1}{c}{EfficientNetB1~\cite{efficientnet}}  & \multicolumn{1}{c}{[A--E,H,I]} & 
\multicolumn{1}{c}{\textbf{90.7\%}}   \\

\multicolumn{1}{c}{EfficientNetB2~\cite{efficientnet}}       & \multicolumn{1}{c}{[A,E]} & 
\multicolumn{1}{c}{90.5\%} \\

\multicolumn{1}{c}{ResNet152~\cite{resnet}}       & \multicolumn{1}{c}{[A,C--E]} & 
\multicolumn{1}{c}{90.3\%} \\

\multicolumn{1}{c}{ResNet152~\cite{resnet}}       & \multicolumn{1}{c}{[A--E,G]} & 
\multicolumn{1}{c}{90.1\%} \\

\multicolumn{1}{c}{EfficientNetB1~\cite{efficientnet}}       & \multicolumn{1}{c}{[A,C--E]} & 
\multicolumn{1}{c}{90.0\%} \\

\multicolumn{1}{c}{ResNet101~\cite{resnet}}       & \multicolumn{1}{c}{[A--E,G,F]} & 
\multicolumn{1}{c}{90.0\%} \\




\hline
\end{tabular}
             \textbf{Notes: }A: ISIC'16 B: ISIC'17 C: ISIC'18 D: ISIC'19 E: ISIC'20 F: 7-point criteria G: PH2 H: PAD\_UFES\_20 I: MEDNODE J: Kaggle

}\end{threeparttable}
}
\end{adjustbox}
\label{table:leaderboard}

}\end{table}

Calibration scores in Table~\ref{table:calibration} show that models using three or more datasets perform better. Notably, shallower models like VGG and EfficientNet often have better calibration than deeper networks such as ResNet and DenseNet, supporting Guo et al. (2017)~\cite{guo2017calibration}. The EfficientNet family exhibits strong stability and accuracy in melanoma detection, as shown in Table \ref{table:leaderboard} and \ref{table:calibration}. Its depthwise convolution with one filter per input channel decreases complexity, enhancing generalization and calibration.
\begin{table}[t!]{
\caption{
Calibration performance of select models based on their classification performance in Table~\ref{table:performance}.}
\centering

\begin{adjustbox}{width=1\columnwidth}{
\begin{threeparttable}{
\normalsize
\begin{tabular}{c c c c c c c}

\hline
\multirow{2}{*}{Testset} & \multirow{2}{*}{Network} & \multirow{2}{*}{Trainsets} & \multicolumn{2}{|c|}{Before rejection} & \multicolumn{2}{c}{After rejection} \\ \multicolumn{3}{c}{} &

\multicolumn{1}{|c}{Brier score ($\downarrow$)} 
& \multicolumn{1}{c}{ECE ($\downarrow$)} &
\multicolumn{1}{|c}{Brier score ($\downarrow$)} 
& \multicolumn{1}{c}{ECE ($\downarrow$)} \\
\hline

\multirow{3}{*}{ISIC2017~\cite{isic2017}} &
\multicolumn{1}{c}{DenseNet201~\cite{densenet}}  & \multicolumn{1}{c}{A--D,H,I} & 
\multicolumn{1}{c}{\textbf{0.0832}} &
\multicolumn{1}{c}{\textbf{0.0158}} &
\multicolumn{1}{c}{\textbf{0.0478}} &
\multicolumn{1}{c}{0.0173} \\

 & \multicolumn{1}{c}{VGG19~\cite{vgg}}       & \multicolumn{1}{c}{[A--E,H,I]} & 
\multicolumn{1}{c}{0.1353} &
\multicolumn{1}{c}{0.0165} &
\multicolumn{1}{c}{0.1140} &
\multicolumn{1}{c}{0.0178} \\

 & \multicolumn{1}{c}{ResNet152~\cite{resnet}}       & \multicolumn{1}{c}{[D]} & 
\multicolumn{1}{c}{0.1353} &
\multicolumn{1}{c}{0.0165} &
\multicolumn{1}{c}{0.1043} &
\multicolumn{1}{c}{\textbf{0.0113}} \\
\hline

\multirow{3}{*}{ISIC2018~\cite{isic2018}} &
\multicolumn{1}{c}{VGG19~\cite{densenet}}  & \multicolumn{1}{c}{A--C,I,J} & 
\multicolumn{1}{c}{\textbf{0.0886}} &
\multicolumn{1}{c}{\textbf{0.0047}} &
\multicolumn{1}{c}{0.0563} &
\multicolumn{1}{c}{\textbf{0.0044}} \\

 & \multicolumn{1}{c}{EfficientNetB1~\cite{efficientnet}}       & \multicolumn{1}{c}{[A,C--E]} & 
\multicolumn{1}{c}{0.1015} &
\multicolumn{1}{c}{0.0246} &
\multicolumn{1}{c}{0.0789} &
\multicolumn{1}{c}{0.0103} \\

 & \multicolumn{1}{c}{VGG16~\cite{vgg}}       & \multicolumn{1}{c}{[A--C,I,J]} & 
\multicolumn{1}{c}{0.0887} &
\multicolumn{1}{c}{0.0131} &
\multicolumn{1}{c}{\textbf{0.0529}} &
\multicolumn{1}{c}{0.0140} \\
\hline

 \multirow{3}{*}{7-point criteria~\cite{7pointcriteria}}
 & \multicolumn{1}{c}{EfficientNetB1~\cite{efficientnet}}       & \multicolumn{1}{c}{[A--E,H,I]} & 
\multicolumn{1}{c}{0.1455} &
\multicolumn{1}{c}{\textbf{0.0288}} &
\multicolumn{1}{c}{0.1218} &
\multicolumn{1}{c}{0.0329} \\

 & \multicolumn{1}{c}{ResNet152~\cite{resnet}}       & \multicolumn{1}{c}{[D]} & 
\multicolumn{1}{c}{\textbf{0.1396}} &
\multicolumn{1}{c}{0.0396} &
\multicolumn{1}{c}{\textbf{0.1119}} &
\multicolumn{1}{c}{\textbf{0.0321}} \\

 & \multicolumn{1}{c}{VGG19~\cite{vgg}}       & \multicolumn{1}{c}{[A--E,G--I]} & 
\multicolumn{1}{c}{0.1428} &
\multicolumn{1}{c}{0.0455} &
\multicolumn{1}{c}{0.1274} &
\multicolumn{1}{c}{0.0458} \\
\hline

 \multirow{3}{*}{Kaggle~\cite{kagglemb}}
 & \multicolumn{1}{c}{VGG16~\cite{vgg}}       & \multicolumn{1}{c}{[A,E,G]} & 
\multicolumn{1}{c}{0.1577} &
\multicolumn{1}{c}{\textbf{0.0187}} &
\multicolumn{1}{c}{0.1357} &
\multicolumn{1}{c}{0.0221} \\

 & \multicolumn{1}{c}{EfficientNetB6~\cite{efficientnet}}       & \multicolumn{1}{c}{[A--C,E,G]} & 
\multicolumn{1}{c}{0.1212} &
\multicolumn{1}{c}{0.0219} &
\multicolumn{1}{c}{0.0919} &
\multicolumn{1}{c}{\textbf{0.0091}} \\

 & \multicolumn{1}{c}{ResNet152~\cite{resnet}}       & \multicolumn{1}{c}{[A--D,I,J]} & 
\multicolumn{1}{c}{\textbf{0.0543}} &
\multicolumn{1}{c}{0.0238} &
\multicolumn{1}{c}{\textbf{0.0200}} &
\multicolumn{1}{c}{0.0122} \\

\hline
\end{tabular}
            \textbf{Notes:} A: ISIC'16 B: ISIC'17 C: ISIC'18 D: ISIC'19 E: ISIC'20 F: 7-point criteria G: PH2 H: PAD\_UFES\_20 I: MEDNODE J: Kaggle

}\end{threeparttable}
}
\end{adjustbox}
\label{table:calibration}

}\end{table}

\noindent
\textbf{Performance After Rejection.} 
ResNet152 improves precision by at least 10\% points (p), representing percentage point differences,
as shown in Table~\ref{table:performance} (after rejection). Other models also demonstrate overall gains. The highest precisions after rejection are 92.8\% (9\%p $\uparrow$), 88.5\% (13.3\%p $\uparrow$), 88.5\% (9.6\%p $\uparrow$), and 94.1\% (0.4\%p $\uparrow$) for benchmarks from ISIC'17, ISIC'18, 7-point criteria, and Kaggle, respectively.
ResNet152 is the only model that consistently improves in ECE, indicating that deeper networks benefit more from rejection techniques.

\begin{figure}[h!]
\centering
\includegraphics[width=0.9\columnwidth]{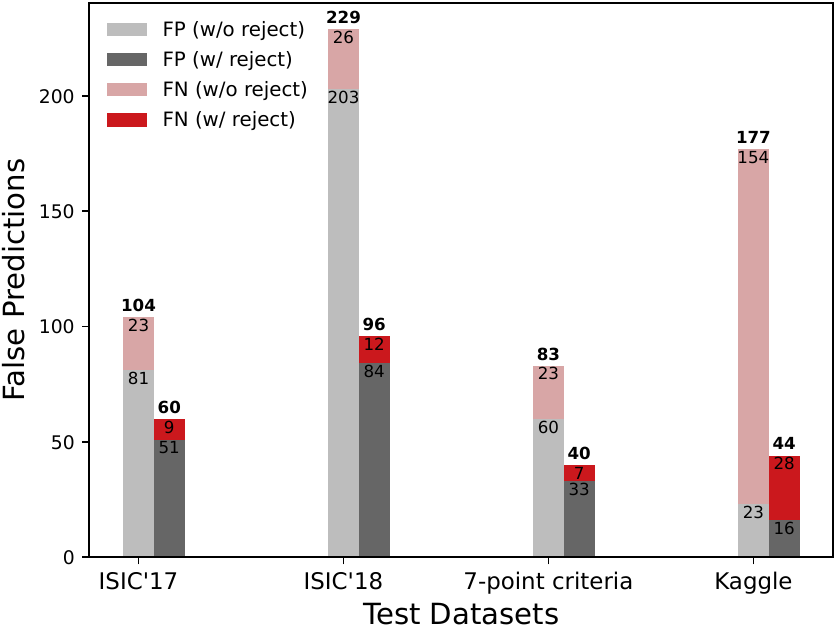}
\caption{A plot comparing false diagnoses before and after applying uncertainty-based rejection across benchmarks.
}
\label{fig:reduced_falsepredictions}
\end{figure}

\noindent
\textbf{Ablation Studies. } 
We conduct additional studies to evaluate our framework's effectiveness. Our findings show a significant reduction in false positives and negatives. Using EfficientNetB1—demonstrated as reliable in Tables~\ref{table:leaderboard} and \ref{table:calibration}—we successfully prevent 353 misdiagnoses, as illustrated in Fig.~\ref{fig:reduced_falsepredictions}. This includes 52\% false positives (non-melanoma diagnosed as melanoma) and 48\% false negatives (melanoma diagnosed as non-melanoma). We reduce 66.5\% of false negatives on average across test sets. Notably, 81\% of false negatives in the Kaggle~\cite{kagglemb} test set are reduced.

\section{Conclusions}
\label{sec:majhead}
We propose a melanoma framework integrating unified training and testing with uncertainty-based rejection to enhance classification and calibration while reducing misdiagnoses. Future plans include collecting real-time diagnosis data from dermatologists and using Generative Adversarial Networks (GAN) for iterative retraining. We also aim to extend the framework to other medical image data, like mammograms for breast cancer detection.

\section{Acknowledgments}

This material is based, in part, upon work supported by the Defense Logistics Agency (DLA) and the Advanced Research Projects Agency for Health (ARPA-H) under Contract Number SP4701-23-C-0073. Any opinions, findings and conclusions or recommendations expressed in this material are those of the author(s) and do not necessarily reflect the views of DLA or ARPA-H. This research was funded in part by the Massachusetts Life Sciences Center through grant Bits-to-Bytes 34428.

\section{Compliance with ethical standards}
\label{sec:print}

All datasets used in this study were anonymized and publicly available, for which no ethical approval was required.

\bibliographystyle{IEEEbib}
\bibliography{strings,refs}

\end{document}